\documentclass{llncs}
\usepackage{paralist}%
\usepackage{import}

\usepackage{ifthen}
\usepackage{url}
\usepackage{amssymb}
\usepackage{amsmath}
\usepackage{import}
\usepackage{xparse}
\usepackage{amsmath}
\usepackage[usenames,dvipsnames]{xcolor}
\usepackage{xspace}
\usepackage{datatool}
\usepackage{glossaries}

\providecommand\m[1]{\ensuremath{#1}\xspace}
\renewcommand{\m}[1]{\ensuremath{#1}\xspace}

	\newcommand{\lrule}{\leftarrow}
	\newcommand{\cause}{\stackrel{c}{\lrule}}

	\newcommand{\voc}{\m{\Sigma}}

	\newcommand{\struct}{\m{I}}

	\newcommand{\PP}{\m{\mathcal{P}}}

	\NewDocumentCommand\inter{g+g}{%
	  \IfNoValueTF{#1}
	    {\struct}
	    {\m{#1^{#2}}}}

	\newcommand{\bool}{\m{\mathbb{B}}}

	\renewcommand{\int}{\m{\mathbb{Z}}}

	\NewDocumentCommand\subs{g+g}{%
	  \IfNoValueTF{#1}
	    {\m{/}}
	    {\m{#1/ #2}}}

	\newcommand{\logicname}[1]{\textsc{#1}\xspace}

	\newcommand{\idp}{\logicname{IDP}}

	\newcommand{\breakid}{\logicname{BreakID}}
	\newcommand{\glucose}{\logicname{Glucose}}
	\newcommand{\shatter}{\logicname{Shatter}}
	\newcommand{\saucy}{\logicname{Saucy}}
	\newcommand{\sbass}{\logicname{sbass}}
	\newcommand{\nauty}{\logicname{nauty}}
	\newcommand{\bliss}{\logicname{bliss}}
	\newcommand{\gringo}{\logicname{gringo}}
	\newcommand{\lparse}{\logicname{Lparse}}
	\newcommand{\smodels}{\logicname{Smodels}}

	\newcommand{\foid}{\logicname{FO(ID)}}

\newcommand{\ouracronym}[3]{%
	\newacronym{#1}{#2}{#3}
	\expandafter\newcommand\csname #1\endcsname{\gls{#1}\xspace}%
}
	\ouracronym{FO}{FO}{first-order logic}
	\ouracronym{PC}{PC}{propositional calculus}
	\ouracronym{MX}{MX}{Model Expansion}
	\ouracronym{MO}{MO}{Model Optimization}
	\ouracronym{ASP}{ASP}{Answer Set Programming}
	\ouracronym{TP}{TP}{Theorem Proving}
	\ouracronym{LP}{LP}{Logic Programming}
	\ouracronym{CP}{CP}{Constraint Programming}
	\ouracronym{FP}{FP}{Functional Programming}
	\ouracronym{KR}{KR}{Knowledge Representation}
	\ouracronym{CSP}{CSP}{Constraint Satisfaction Problem}
	\ouracronym{SMT}{SMT}{SAT Modulo Theories}
	\ouracronym{KBS}{KBS}{Knowledge Base System}
	\ouracronym{NNF}{NNF}{Negation Normal Form}
	\ouracronym{FNNF}{FNNF}{Flat Negation Normal Form}
	\ouracronym{DefNNF}{DefNNF}{Definition Negation Normal Form}
	\ouracronym{DEFNF}{DEFNF}{Definition Normal Form}
	\ouracronym{CDCL}{CDCL}{Conflict-Driven Clause-Learning}
	\ouracronym{WFS}{WFS}{Well-Founded Semantics}
	\ouracronym{LCG}{LCG}{Lazy Clause Generation}
	\ouracronym{AEL}{AEL}{Autoepistemic Logic}
	\ouracronym{OEL}{OEL}{Ordered Epistemic Logic}
	\ouracronym{AFT}{AFT}{Approximation Fixpoint Theory}

	\makeatletter
	\def\ifenv#1{
	\def\@tempa{#1}%
	\def\@ttempa{#1*}%
	\ifx\@tempa\@currenvir
	\expandafter\@firstoftwo
	\else
	\expandafter\@secondoftwo
	\fi
	}
	\makeatother

	\newcommand{\ddrule}[4]{\ensuremath{#1 \leftarrow #2 & \{#3\} & #4}}
	\newcommand{\drule}[2]{\ensuremath{#1 & \leftarrow & #2}}

	\newcommand{\darule}[4]{\ensuremath{#1 \leftarrow #2 & \{#3\} & #4}}
	\newcommand{\arule}[2]{\ensuremath{#1 \, &\leftarrow \, #2}}

	\newcommand{\LNDRule}[2]{
	\ifenv{array}
	{\drule{#1}{#2}}
	{ \ifenv{align}
		{\arule{#1}{#2}}
		{\ifenv{align*}
		{\arule{#1}{#2}}
		{ERROR: using LDRule in unsupported environment: \@currenvir}
		}
	}
	}

	\newcommand{\LDRule}[4]{
	\ifenv{array}
	{\ddrule{#1}{#2}{#3}{#4}}
	{ \ifenv{align}
		{\darule{#1}{#2}{#3}{#4}}
		{\ifenv{align*}
		{\darule{#1}{#2}{#3}{#4}}
		{ERROR: using LDRule in unsupported environment: \@currenvir}
		}
	}
	}

	\NewDocumentCommand\LRule{m+g+g+g}{%
		\IfNoValueTF{#2}%
		{#1.&}{%
		\IfNoValueTF{#3}
		{\LNDRule{#1}{#2.}}
		{\LDRule{#1}{#2.}{#3}{#4}}%
		}
	}

	\NewDocumentCommand\CLRule{m+g}{%
	\ifenv{array}
	{\cdrule{#1}{#2}}
	{ \ifenv{align}
		{\carule{#1}{#2}}
		{\ifenv{align*}
			{\carule{#1}{#2}}
			{ERROR: using CLRule in unsupported environment: \@currenvir}
		}
	}
	}

	\NewDocumentCommand\carule{m+g}{%
		\IfNoValueTF{#2}
			{\ensuremath{#1.}}
			{\ensuremath{#1 \, &\cause \, #2}}}
	\NewDocumentCommand\cdrule{m+g}{%
		\IfNoValueTF{#2}
			{\ensuremath{#1.}}
			{\ensuremath{#1 & \cause & #2}}}

	\newcommand{\algrule}[4]{
	\hbox{{#1}:}& 
	\quad #2 ~\longrightarrow~ #3 
	\hbox{~ if } #4\\
	}

	\newcommand{\AlgoRule}[4]{
	\ifenv{array}
	{\algrule{#1}{#2}{#3}{#4}}
		{ERROR: using AlgoRule in unsupported environment: \@currenvir}
	}

\newcommand{\commentstyle}{\color{Gray}}

	\usepackage[final]{listings}

	\lstdefinelanguage{idp}{
		morekeywords=[1]{query(}, 
		morekeywords=[2]{namespace,vocabulary,theory,structure,procedure,term,set,formula, spec, specification,query},
		morekeywords=[3]{include,using,type,isa,contains,partial,extern,LFD,GFD,constructed,from,constraint,pred,supertype,of,subtype,define},
		morekeywords=[4]{int,float,char,string,nat},
		morekeywords=[5]{if,then,else,for,end},
		morecomment=[s]{/*}{*/},	
		morecomment=[l]{//}
	}
	\newcommand{\ignore}[1]{}

	\newboolean{nocomments}
	\setboolean{nocomments}{false}

	\newboolean{commentmargin}
	\setboolean{commentmargin}{true}

	\newcommand{\namedcomment}[3]{%
		\ifthenelse{\boolean{nocomments}}%
		{}
		{
			\ifthenelse{\boolean{commentmargin}}%
				{ {\color{#3} \marginpar{\color{#3}\sc #2}#1}  }
				{  {\color{#3} {\sc #2}: #1}  }
		}%
	}
	\newcommand{\mnamedcomment}[3]{\ifthenelse{\boolean{nocomments}}{}{{\marginpar{ \color{#3}{\sc #2}:#1}}}}

	\usepackage{soul}

\usepackage{etoolbox}

\newcommand\setcitation[2]{%
  \csdef{mycommoncitation#1}{#2}}
\newcommand\getcitation[1]{%
  \csuse{mycommoncitation#1}}

\setcitation{IDP}{WarrenBook/DeCatBBD14}
\setcitation{idp}{WarrenBook/DeCatBBD14}
\setcitation{fodot}{tocl/DeneckerT08}
\setcitation{CP}{Apt03}
\setcitation{cp}{Apt03}
\setcitation{EZCSP}{lpnmr/Balduccini11}
\setcitation{KR}{Baral:2003}
\setcitation{ASPComp2}{lpnmr/DeneckerVBGT09}
\setcitation{ASPComp3}{journals/tplp/CalimeriIR14}
\setcitation{ASPComp4}{conf/lpnmr/AlvianoCCDDIKKOPPRRSSSWX13}
\setcitation{CPSupport}{ictai/DeCat13}
\setcitation{CPsupport}{ictai/DeCat13}
\setcitation{functionDetection}{iclp/DeCatB13}
\setcitation{fodot2asp}{DeneckerLTV12} 
\setcitation{Tarskian}{DeneckerLTV12} 
\setcitation{Inca}{iclp/DrescherW12}
\setcitation{csp2asp}{ijcai/DrescherW11}
\setcitation{DPLLT}{cav/GanzingerHNOT04}
\setcitation{AspInPractice}{synthesis/2012Gebser}
\setcitation{clasp}{ai/GebserKS12}
\setcitation{oclingo}{kr/GebserGKOSS12}
\setcitation{gringo}{lpnmr/GebserST07}
\setcitation{cmodels}{aaai/GiunchigliaLM04}
\setcitation{inputster}{tplp/Jansen13}
\setcitation{DLV}{tocl/LeonePFEGPS06}
\setcitation{LearningPaper}{TPLP/BruynoogheBBDDJLRDV} 
\setcitation{clog}{iclp/BogaertsVDV14} 
\setcitation{foc}{iclp/BogaertsVDV14}
\setcitation{inferenceClog}{ecai/BogaertsVDV14}
\setcitation{examplesClog}{nmr/BogaertsVDV14b} 
\setcitation{AFT}{DeneckerMT00}
\setcitation{KBS}{iclp/DeneckerV08}
\setcitation{KBPE}{inap/DePooterWD11}
\setcitation{lazyGrounding}{jair/CatDBS15} 
\setcitation{lazygrounding}{jair/CatDBS15} 
\setcitation{ASP}{marek99stable}
\setcitation{satid}{sat/MarienWDB08}
\setcitation{lazyclausegeneration}{constraints/OhrimenkoSC09}
\setcitation{FP}{ACMCS/Hudak89}
\setcitation{GroundingWithBounds}{jair/WittocxMD10}
\setcitation{SAT}{faia/SilvaLM09}
\setcitation{LTC}{iclp/Bogaerts14}
\setcitation{SPSAT}{ictai/DevriendtBMDD12}
\setcitation{BreakID}{sat/DevriendtBBD16}
\setcitation{breakid}{sat/DevriendtBBD16}
\setcitation{LCG}{stuckeyLCG}
\setcitation{MiniZinc}{conf/cp/NethercoteSBBDT07}
\setcitation{minizinc}{conf/cp/NethercoteSBBDT07}
\setcitation{amadini}{cpaior/AmadiniGM13}
\setcitation{bootstrapping}{lash/BogaertsJDJBD14}
\setcitation{GroundedFixpoints}{ai/BogaertsVD15}
\setcitation{LogicBlox}{datalog/GreenAK12}
\setcitation{proB}{journals/sttt/LeuschelB08}
\setcitation{NaturalInductions}{KR/DeneckerV14} 
\setcitation{LP}{jacm/EmdenK76}
\setcitation{SMT}{faia/BarrettSST09}
\setcitation{AF}{ai/Dung95}
\setcitation{ADF}{kr/BrewkaW10}
\setcitation{af}{ai/Dung95}
\setcitation{adf}{kr/BrewkaW10}
\setcitation{ADFRevisited}{ijcai/BrewkaSEWW13}
\setcitation{adfrevisited}{ijcai/BrewkaSEWW13}
\setcitation{DefaultLogic}{ai/Reiter80}
\setcitation{DL}{ai/Reiter80}
\setcitation{AEL}{mo85}
\setcitation{minisat}{sat/EenS03}
\setcitation{completion}{adbt/Clark78}
\setcitation{wasp}{lpnmr/AlvianoDFLR13}
\setcitation{minisatid}{ictai/DeCat13}
\setcitation{lcg}{stuckeyLCG}
\setcitation{CEGAR}{jacm/ClarkeGJLV03}
\setcitation{cegar}{jacm/ClarkeGJLV03}
\setcitation{CuttingPlane}{or/DantzigFJ54}
\setcitation{kodkod}{tacas/TorlakJ07}
\setcitation{cdcl}{Marques-SilvaS99}
\setcitation{CDCL}{Marques-SilvaS99}
\setcitation{relevance}{ijcai/JansenBDJD16}
\setcitation{WFS}{GelderRS91}
\setcitation{wfs}{GelderRS91}
\setcitation{UnfoundedSet}{GelderRS91}
\setcitation{stablesemantics}{iclp/GelfondL88}
\setcitation{StableSemantics}{iclp/GelfondL88}
\setcitation{shatter}{Shatter}
\setcitation{sbass}{drtiwa11a}
\setcitation{lparsemanual}{url:lparse_manual}
\setcitation{}{}
\setcitation{}{}
\setcitation{}{}
\setcitation{}{}
\setcitation{}{}
\setcitation{}{}
\setcitation{}{}
\setcitation{}{}
\setcitation{}{}
\setcitation{}{}
\setcitation{}{}
\setcitation{}{}
\setcitation{}{}
\setcitation{}{}
\setcitation{}{}
\setcitation{}{}
\setcitation{}{}
\setcitation{}{}
\setcitation{}{}
\setcitation{}{}
\setcitation{}{}
\setcitation{}{}
\setcitation{}{}
\setcitation{}{}
\setcitation{}{}
\setcitation{}{}
\setcitation{}{}
\setcitation{}{}
\setcitation{}{}
\setcitation{}{}
  
\newcommand\refto[1]{%
      \getcitation{#1}}
      
\newcommand\mycite[1]{%
      \ifcsname mycommoncitation#1\endcsname%
   \cite{\getcitation{#1}}%
  \else%
    \cite{#1}%
  \fi%
}	
  
\ignore{

}
\newcommand{\cominisatps}{\logicname{COMiniSatPS}}
\usepackage{tikz}
\usetikzlibrary{arrows,positioning,fit,shapes,calc}
\usetikzlibrary{shadows}

\newcommand\clasp{\logicname{clasp}}

\newcommand\graceful{\textbf{graceful}}
\newcommand\holes{\textbf{pigeons}}
\newcommand\nqueens{\textbf{queens}}
\newcommand\crew{\textbf{crew}}
\newcommand\valves{\textbf{valves}}
\newcommand\still{\textbf{still}}

\setboolean{nocomments}{false}

\begin{document}
\frontmatter          
\pagestyle{headings}  
\mainmatter              
\title{BreakID: Static Symmetry Breaking for ASP (System Description)}
\titlerunning{BreakID: Static Symmetry Breaking for ASP}  
%
\author{Jo Devriendt\inst{1} \and Bart Bogaerts\inst{2,1}}
\authorrunning{Devriendt and Bogaerts} 
%
%
\institute{KU Leuven, Department of Computer Science\\ Celestijnenlaan 200A, Leuven, Belgium\\
\email{firstname.lastname@cs.kuleuven.be}
\and
Helsinki Institute for Information Technology HIIT, Aalto University\\ FI-00076 AALTO, Finland}

\maketitle              

\begin{abstract}
Symmetry breaking has been proven to be an efficient preprocessing technique for satisfiability solving (SAT). 
In this paper, we port the state-of-the-art SAT symmetry breaker \breakid to answer set programming (ASP). 
The result is a lightweight tool that can be plugged in between the grounding and the solving phases that are common when modelling in ASP. 
We compare our tool with \sbass, the current state-of-the-art symmetry breaker for ASP.
\end{abstract}
\section{Introduction}
Answer set programming (ASP) has always benefited from progress in the satisfiability solving (SAT) community. 
In fact, many (if not all) modern ASP solvers \cite{\refto{clasp},\refto{wasp},\refto{minisatid}} are based on conflict-driven clause learning (CDCL) \mycite{cdcl}, a technique first developed for SAT. 
Some people even go as far as to claim that ASP is just ``a good modelling language for SAT''.

Thus, now and then it makes sense to check which developments in SAT can be transferred to ASP. 
In this paper, we port a recently introduced symmetry breaking preprocessor, \breakid\footnote{Pronounced ``break it''.} \mycite{breakid} to ASP. 
A first version of \breakid was developed in 2013 \cite{cspsat/DevriendtBB14}. 
This version won a gold medal on the 2013 SAT competition \cite{url:SATcomp2013} in combination with the SAT solver  \glucose 2.1 \mycite{glucose}.
The latest version of \breakid was developed in 2015 \mycite{breakid} and won the gold medal in the \emph{no limit} track of the 2016 SAT competition \cite{url:SATcomp2016} in combination with the SAT solver \cominisatps~\cite{phd/Oh15}.

Symmetry breaking has been around for quite a long time. 
The most well-known symmetry breaker for SAT is \shatter \mycite{shatter}. 
\shatter works by \textit{(i)} transforming a SAT theory into a colored graph whose automorphisms correspond to symmetries of the theory,  \textit{(ii)} calling a graph automorphism detection tool such as \saucy \mycite{saucy} to obtain a set of generators of the symmetry group, and  \textit{(iii)} adding constraints that break each of the generators returned by the previous step. 
\breakid is based on the same ideas, but improves the last step of the symmetry breaking process.
In particular, the set of symmetry breaking constraints is parametrized by both the chosen set of generator symmetries, as well as a chosen order on the set of propositional variables.
Good choices for these two parameters strongly influence the breaking power and the size of the resulting constraints \mycite{breakid} .
As such, \breakid 
\begin{compactitem}
 \item searches for more structure in the symmetry breaking group, essentially detecting subgroups of the symmetry group that can be broken completely without additional overhead,
 \item reduces the size of symmetry breaking constraints,
 \item chooses an order on the variables that ``matches'' the generators to break,
 \item actively searches for small (size 2) symmetry breaking clauses.
\end{compactitem}
This resulted in significant improvements with respect to \shatter. 

The techniques from \shatter have been ported to ASP by the ASP symmetry breaker \sbass \mycite{sbass}. 
The main contribution there was to develop an alternative for the first step in the ASP context, i.e., to transform an ASP program \PP into a colored graph such that graph isomorphisms correspond to symmetries of \PP. 

In this paper, we show how to use \breakid as a symmetry breaker for ASP. 
Our implementation extends (and slightly modifies) the graph encoding used in \sbass. 
The extensions serve to support a richer language: we provide support for so-called \emph{weight rules} and \emph{minimize statements}. 
The modification serves to ensure more effective breaking; we present several simple symmetric examples where \breakid detects the symmetry while \sbass does not.
For the third step (generating symmetry breaking constraints), we use all of the 
improvements that are already present in the SAT-version of \breakid~\mycite{breakid}. 

We evaluate performance of \breakid and \sbass on a number of benchmarks with symmetries and conclude that also in an ASP context, \breakid's improvements have a significant effect. 


The rest of this paper is structured as follows. In Section \ref{sec:prelims} we recall preliminaries regarding answer set programming and symmetries. Afterwards, in Section \ref{sec:breaking} we present our symmetry breaker for ASP. We evaluate this tool in Section \ref{sec:exp} and conclude in Section \ref{sec:concl}.

\section{Preliminaries}\label{sec:prelims}
\newcommand\head{\m{\mathit{head}}}
\newcommand\body{\m{\mathit{body}}}
\subsection{Answer Set Programming}


A \emph{vocabulary} is a set of symbols, also called
\emph{atoms}; vocabularies are denoted by $\sigma,\tau$. A \emph{literal}
is an atom or its negation. An interpretation $\struct$ of the vocabulary \voc is a subset of $\voc$.
 Propositional formulas and the satisfaction relation between formulas and interpretations are defined as usual.
A \emph{logic program} $\PP$ over vocabulary $\voc$ is a set of \emph{rules} $r$ of form
\begin{equation}\label{eq:rule} 
h_1\lor \dots \lor h_l \lrule a_1\land \dots \land a_n \land \lnot b_1\land \dots \land \lnot b_m,\end{equation}
where $h_i$'s, $a_i$'s, and $b_i$'s are atoms in $\voc$. The formula $h_1\lor \dots \lor h_l$ is called the \emph{head} of $r$, denoted $\head(r)$, and the formula 
$a_1\land \dots \land a_n \land \lnot b_1\land \dots \land \lnot b_m$ the \emph{body} of $r$, denoted $\body(r)$. A program is called \emph{normal} (resp. \emph{positive}) if $l=1$ (resp. $m=0$) for all rules in \PP. If $n=m=0$, we simply write $h_1\lor \dots \lor h_l$. If $l=0$, we call $r$ a \emph{constraint}.

An interpretation $I$ is a \emph{model} of a logic program \PP if, for all rules $r$ in
\PP, whenever $\body(r)$ is satisfied by $I$, so is $\head(r)$.
The \emph{reduct} of \PP with respect to $I$, denoted $\PP^I$, is a positive program that consists of rules 
$
h_1\lor \dots \lor h_l \lrule a_1\land \dots \land a_n $
for all rules of the form \eqref{eq:rule} in \PP such that $b_i\not\in I$ for all $i$. 
%
An interpretation $I$ is a \emph{stable model} or an \emph{answer set} of \PP if it is a $\subseteq$-minimal model of $\PP^I$ \cite{iclp/GelfondL88}.
Checking whether a logic program has a stable model is an $\Sigma_2^P$-complete task in general and an NP-complete task for normal programs; hence, logic programs can be used to encode search problems.
This observation gave birth to the field of answer set programming \cite{marek99stable,Niemela99,iclp/Lifschitz99}.


This paper sometimes uses syntactic sugar such as choice rules, constraints,
cardinality atoms and weight atoms. A \emph{cardinality atom} \[g\leq \#\{l_1, \dots, l_f\} \leq k\] (with $l_1, \dots,
l_f$ being literals and $f,g, k \in \mathbb{N}$) is satisfied by
$I$ if
\[g \leq \#\{i\mid I \models l_i\} \leq k.\]
A \emph{weight atom} \[g\leq \mathrm{sum}\{l_1=w_i, \dots, l_f=w_f\} \leq k\] (with $l_1, \dots,
l_f$ being literals and $f, g, k, w_i \in \mathbb{N}$) is satisfied if 
\[g \leq \sum_{\{i \mid I \models l_i\}} w_i \leq k.\]
Sometimes, the ``$g\leq$'' or ``$\leq k$'' parts of cardinality or weight atoms are dropped. In this case, the obvious semantics applies: the atom is satisfied for any lower/upper bound. For instance, the cardinality atom 
\[g\leq \#\{l_1, \dots, l_f\} \]
is satisfied whenever
\[g \leq \#\{i\mid I \models l_i\} .\]
A \emph{choice rule} is a rule
with a cardinality atom in the head, i.e., a rule of the form
\[g \leq \#\{l_1, \dots, l_f\} \leq k\lrule a_1\land \dots \land a_n \land \lnot b_1\land \dots \land \lnot b_m.\]
$I$ satisfies a constraint $c$ if it does not satisfy
$body(c)$. These language constructs can all be translated to normal rules \cite{lpnmr/BomansonJ13}.

\subsection{Symmetry}
 Let $\pi$ be a permutation of $\voc$. We extend $\pi$ to literals: $\pi(\neg a)=\neg(\pi(a))$, to rules:
 \begin{align*}
 &\pi(h_1\lor \dots \lor h_l \lrule a_1\land \dots \land a_n \land \lnot b_1\land \dots \land \lnot b_m)= \\
 & \quad \pi(h_1)\lor \dots \lor \pi(h_l) \lrule \pi(a_1)\land \dots \land \pi(a_n) \land \lnot \pi(b_1)\land \dots \land \lnot \pi(b_m),
 \end{align*}
 to
 logic programs: $\pi(\PP)=\{\pi(r)\mid r\in \PP\}$, and to interpretations: $\pi(\struct)=\{\pi(p)\mid p\in \struct\}$. 
A \emph{symmetry} of a program $\PP$ is a permutation $\pi$ of $\voc$
that \emph{preserves stable models} of $\PP$ (i.e., $\pi(\struct)$ is a stable model of $\PP$ if and only if  $\struct$ is a stable model of $\PP$).
A permutation of atoms for which $\pi(\PP)=\PP$ is a \emph{syntactical} symmetry of $\PP$.\footnote{In some texts, the term \emph{symmetry} is reserved for what we call \emph{syntactical symmetry}.}
Typically, only syntactical symmetry is exploited, since this type of symmetry can be detected with relative ease. 
The practical techniques presented in this paper are no exception. 

\emph{Symmetry breaking} aims at eliminating symmetry, either by
\emph{statically} posting a set of constraints that invalidate
symmetric interpretations, or by altering the search space \emph{dynamically} to avoid symmetric search paths. 
In this paper, we focus on static symmetry breaking. 
If $\Pi$ is a symmetry group, then a set of constraints $\psi$ is \emph{sound} if for each interpretation $\struct$ there exists at least one symmetry $\pi \in \Pi$ such that $\pi(\struct)$ satisfies $\psi$;
$\psi$ is \emph{complete} if for each assignment $\alpha$ there exists at most one symmetry $\pi \in \Pi$ such that $\pi(\alpha)$ satisfies $\psi$~\cite{walsh_recent_results_2012}.

\subsection{Graph Automorphisms}
A \emph{colored graph} is a tuple $G=(V,E,c)$, where $V$ is a set, whose elements we call \emph{nodes}, $E$ is a binary relation on $V$; elements of $E$ are called edges and $c$ is a mapping $V\to C$ for some set $C$. The elements of $C$ are called \emph{colors}.    

An automorphism of $G$ is a mapping $\pi: V\to V$ such that the following two conditions hold:
\begin{compactitem}
 \item $(u,v)\in E$ if and only if $(\pi(u),\pi(v))\in E$ for each $u,v \in V$, and
 \item $c(v) = c(\pi(v))$ for each $v\in V$.
\end{compactitem}

The graph automorphism problem is the task of finding all graph automorphisms of a given (colored) graph. 
The complexity of this problem is conjectured to be strictly in between P and NP \cite{Babai95}. 
Several tools are available to tackle this problem, including \saucy \mycite{saucy}, \nauty \mycite{nauty} and \bliss \mycite{bliss}. 


\section{Symmetry Breaking for ASP}\label{sec:breaking}
As mentioned in the introduction, symmetry breaking typically consists of three steps. 
\begin{compactitem}
 \item First, a colored graph $G$ is created such that the automorphisms of $G$ correspond to symmetries of \PP. 
 \item Second, a set $\mathcal{G}$ of generators of the automorphism group of $G$ is derived. 
 \item Third, a set of symmetry breaking constraints is added to the original problem, based on $\mathcal{G}$.
\end{compactitem}

As for the second step, any graph automorphism tool can be used.
\shatter \mycite{shatter}, \sbass and  \breakid make use of \saucy for this task. 
We discuss how we tackle the first and last step in the following subsections. 

\subsection{Colored Graph Encoding}
\begin{figure}
\centering
\begin{tikzpicture}[node distance=2cm, auto,]

 \node[rounded corners, draw=black,drop shadow,minimum size=0.75cm,fill=blue!30] at (0, -0.5) (h1) {$h_l$};
 \node[rounded corners, draw=black,drop shadow,minimum size=0.75cm,fill=blue!30] at (0, 1) (hdots) {$\dots$};
 \node[rounded corners, draw=black,drop shadow,minimum size=0.75cm,fill=blue!30] at (0, 2.5) (hl) {$h_1$};

  \node[rounded corners, draw=black,drop shadow,minimum size=0.75cm,fill=red!50] at (2, 1) (head) {$\head(r)$};
  \node[rounded corners, draw=black,drop shadow,minimum size=0.75cm,fill=orange!50] at (4, 1) (body) {$\body(r)$};
  
   \node[rounded corners, draw=black,drop shadow,minimum size=0.75cm,fill=blue!30] at (5, -0.5) (a1) {$a_1$};
 \node[rounded corners, draw=black,drop shadow,minimum size=0.75cm,fill=blue!30] at (6, 0) (adots) {$\dots$};
 \node[rounded corners, draw=black,drop shadow,minimum size=0.75cm,fill=blue!30] at (7, 0.5) (an) {$a_n$};

   \node[rounded corners, draw=black,drop shadow,minimum size=0.75cm,fill=green!50] at (5,2.5) (b1) {$\lnot b_1$};
 \node[rounded corners, draw=black,drop shadow,minimum size=0.75cm,fill=green!50] at (6, 2) (bdots) {$\dots$};
 \node[rounded corners, draw=black,drop shadow,minimum size=0.75cm,fill=green!50] at (7, 1.5) (bm) {$\lnot b_m$};
 
 \draw (h1)--(head);
 \draw (hdots)--(head);
 \draw (hl)--(head);
  \draw (a1)--(body);
 \draw (adots)--(body);
 \draw (an)--(body);
  \draw (b1)--(body);
 \draw (bdots)--(body);
 \draw (bm)--(body);
 \draw (head)--(body);

 \end{tikzpicture}
\caption{Encoding of a rule $r$ of the form \eqref{eq:rule}.}\label{fig:rule}
\end{figure}
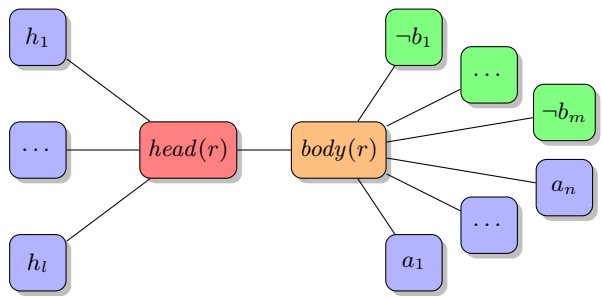

As for encoding a logic program as a colored graph, our approach is very close to the one introduced by \sbass \mycite{sbass} (we discuss and justify the differences below). 
We use four colors $\{1,\dots,4\}$; our graph consists of the following nodes:
\begin{compactitem}
 \item For each atom $p\in \voc$, two nodes, referred to as $p$ and $\lnot p$ below. Node $p$ is colored as $1$, node $\lnot p$ is colored $2$. 
 \item For each rule $r\in \PP$, two nodes, referred to as $\head(r)$ and $\body(r)$ below.  Node $\head(r)$ is colored $3$ and node $\body(r)$ is colored $4$. 
\end{compactitem}
Our graph is \emph{undirected} and the edges are as follows:
\begin{compactitem}
 \item Each node $p$ is connected to $\lnot p$. 
 \item For each rule $r$ of the form \eqref{eq:rule} in \PP, each node $h_i$ is connected to $\head(r)$, each node $a_i$ is connected to $\body(r)$ and each node $\lnot b_i$ is connected to $\body(r)$. The complete encoding of this type of  rule is illustrated in Figure~\ref{fig:rule}.
\end{compactitem}

It can be seen that there is a one-to-one correspondence between automorphisms of this graph and syntactic symmetries of the logic program. Since atoms  $p$ are the only nodes colored in color $1$, an automorphism  induces a permutation of \voc. The edge between $p$ and $\lnot p$ guarantee that an automorphism that maps to $p$ to $q$ also maps $\lnot p$ to $\lnot q$. 
Furthermore, the edges between $\head(r), \body(r)$ and the various literals occurring in a rule capture the full structure of the rule. As such, it can be verified that automorphisms of this graph must map rules to ``syntactically symmetric'' rules. Vice versa, each syntactic symmetry preserves the graph structure defined above. 

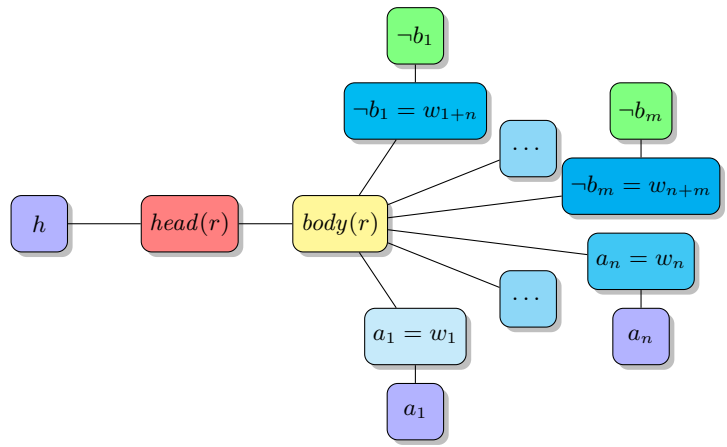
\begin{figure}
\centering
\begin{tikzpicture}[node distance=2cm, auto,]

 \node[rounded corners, draw=black,drop shadow,minimum size=0.75cm,fill=blue!30] at (0, 1) (h) {$h$};

  \node[rounded corners, draw=black,drop shadow,minimum size=0.75cm,fill=red!50] at (2, 1) (head) {$\head(r)$};
  \node[rounded corners, draw=black,drop shadow,minimum size=0.75cm,fill=yellow!50] at (4, 1) (body) {$\body(r)$};
  
   \node[rounded corners, draw=black,drop shadow,minimum size=0.75cm,fill=cyan!20]  at (5, -0.5)  (a1) {$a_1=w_1$};
 \node[rounded corners, draw=black,drop shadow,minimum size=0.75cm,fill=cyan!40] at (6.5, 0)(adots) {$\dots$};
 \node[rounded corners, draw=black,drop shadow,minimum size=0.75cm,fill=cyan!60] at (8, 0.5) (an) {$a_n=w_n$};

   \node[rounded corners, draw=black,drop shadow,minimum size=0.75cm,fill=cyan!80] at(5,2.5) (b1) {$\lnot b_1=w_{1+n}$};
 \node[rounded corners, draw=black,drop shadow,minimum size=0.75cm,fill=cyan!40] at (6.5, 2) (bdots) {$\dots$};
 \node[rounded corners, draw=black,drop shadow,minimum size=0.75cm,fill=cyan] at (8, 1.5) (bm) {$\lnot b_m=w_{n+m}$};
 
 \draw (h)--(head);
  \draw (a1)--(body);
 \draw (adots)--(body);
 \draw (an)--(body);
  \draw (b1)--(body);
 \draw (bdots)--(body);
 \draw (bm)--(body);
 \draw (head)--(body);

   \node[rounded corners, draw=black,drop shadow,minimum size=0.75cm,fill=blue!30]  at (5, -1.5)  (aa1) {$a_1$};
   
      \node[rounded corners, draw=black,drop shadow,minimum size=0.75cm,fill=blue!30]  at (8, -0.5)  (aan) {$a_n$};
   
      \node[rounded corners, draw=black,drop shadow,minimum size=0.75cm,fill=green!50]  at (5, 3.5)  (bb1) {$\lnot b_1$};
   
      \node[rounded corners, draw=black,drop shadow,minimum size=0.75cm,fill=green!50]  at (8, 2.5)  (bbm) {$\lnot b_m$};
   
     \draw (a1)--(aa1);
     \draw (an)--(aan);
     \draw (b1)--(bb1);
     \draw (bm)--(bbm);

 \end{tikzpicture}
\caption{Encoding of a rule $r$ of the form \eqref{eq:weight}. The color of $\body(r)$ depends on the bound $g$. The color of each $l_j=w_i$ depends on the value $w_i$.}\label{fig:weight}
\end{figure}

Additionally, we extend this graph encoding to capture more language extensions such as, e.g., \emph{choice rules}. We only support limited forms of these rules. Actually, our tool supports \emph{exactly} the \lparse-\smodels intermediate format \mycite{lparsemanual}, extended with support for disjuctive rules.
This means for instance that in cardinality rules, we assume there is no upper. The general language can easily be transformed to this format (in fact, that is what all modern ASP grounders do). 
For this, we introduce two new colors, extending the set of colors to $\{1,\ldots,6\}$.
In programs with these extensions, integer numbers can occur, either as bounds of cardinality or weight constraint or as weights in a weight constraint or minimize statement (see below). 
We assume that for each integer $n$ that occurs in such a program, there is a unique color $c_n\not\in\{1,\dots,6\}$ available and extend our set of colors to 
\[\{1,\dots,6\}\cup\{c_n\mid n \text{ occurs as weight or bound in \PP}\}.\]
\begin{description}
 \item[Cardinality rules] Rules of the form 
 \[h\lrule g \leq \#\{a_1, \dots , a_n , \lnot b_1, \dots , \lnot b_m\}\]
 are encoded as follows. 
 The head of the rule is encoded as usual. The body node of the rule is colored in $c_g$, the color associated with the bound $g$. 
 The body node is connected to each of the $a_i$ and $\lnot b_i$ and to the head node, as usual. 
 \item[Choice rules]Rules of the form 
 \[\#\{h_1, \dots, h_l\}   \lrule a_1\land \dots \land a_n \land \lnot b_1\land \dots \land \lnot b_m \]
 are encoded exactly the same as rules of form \eqref{eq:rule}, except that $\head(r)$ is colored $5$. This allows to differentiate between standard rules and choice rules. 
 \item[Weight rules] Rules of the form 
 \begin{equation}h\lrule g \leq \mathrm{sum}\{a_1=w_1, \dots, a_n=w_n, \lnot b_1=w_{n+1}, \dots, \lnot b_m=w_{n+m}\}\label{eq:weight}\end{equation}
 are encoded as follows. The node $\body(r)$ is colored as for cardinality rules in $c_g$. 
 For each occurrence of 
 an expression $l_i=w_j$, we create one additional node, referred to as $l_i=w_j$. This node is colored in $c_{w_j}$, the color associated to the integer $w_j$ and 
 is connected to $l_i$.  The body of this rule is connected to all the nodes $a_i=w_i$ and $\lnot b_i=w_{n+i}$. $\body(r)$ is connected to $\head(r)$ and $\head(r)$ to $h$, as usual. A visualisation of this encoding can be found in Figure \ref{fig:weight}. 
 \item[Minimize statements] Minimize statements are expressions of the form 
 \begin{equation}\label{eq:minimize}\mathrm{minimize} \{a_1=w_1, \dots, a_n=w_n, \lnot b_1=w_{n+1}, \dots, \lnot b_m=w_{n+m}\}\end{equation}
 They are directions to the solver that the user is only interested in models such that the term 
 \[\sum_{\text{\{rules of the form \eqref{eq:minimize} that occur in \PP\}}} \sum_{\{i \mid I \models l_i\}} w_i.\]
 is minimal (among all stable models). Such a statement is encoded analogously to the body of a weight rule, except that the body node is replaced by a \emph{minimize node} with color $6$. 
 \end{description}

\paragraph{Comparison with \sbass}
Our graph encoding differs from the one used by \sbass in two respects. 

First of all, we added support for \emph{minimize statements} and \emph{weight rules}. 
As a result, our tool supports the full \lparse-\smodels intermediate language as documented in \mycite{lparsemanual} and additionally, the rule type ``8'' used by \gringo \mycite{gringo} to represent disjunctive rules. 

Second, we use \emph{undirected graphs} whereas \sbass uses directed graphs.
To be precise, when using directed graphs, all $\head(r)$ nodes can be dropped by using edges from literals $a_i$ and $\lnot b_i$ to $\body(r)$ and from $\body(r)$ to $h_i$ (the directionality thus distinguishes between head and body literals). We expect because of this that symmetry detection in \breakid takes slightly more time than symmetry detection in \sbass (automorphism algorithms are sensitive to the number of nodes in the input graph). 
The reason why we did this is that we experimentally noticed that \saucy does not always behave well with directed graphs. On very small examples already, symmetries are missed when using the directional encoding. \sbass, which uses \saucy with directed graphs, shows the same kind of behaviour, as we show in the following list of examples. The examples illustrate that \breakid is more stable than \sbass with respect to symmetry detection on different encodings. 

\begin{example}\label{ex:choice}
 \[\PP_1=\left\{ \begin{array}{l}
               0\leq \#\{p\} \leq 1.\\
               0\leq \#\{q\} \leq 1.
              \end{array}
 \right\}.\]
  It is clear that $p$ and $q$ are interchangeable in $\PP_1$. By this we mean that the mapping $\sigma \colon \{p,q\}\to\{p,q\} \colon p\mapsto q, q\mapsto p$ is a symmetry of $\PP_1$.
 \breakid detects this, while \sbass detects (and breaks) no symmetry. 
 \end{example}
 
 \begin{example}
 
 Consider the logic program 
  \[\PP_2=\left\{ \begin{array}{l}
               r\lrule p\land q. \\
               0\leq \#\{p\} \leq 1.\\
               0\leq \#\{q\} \leq 1.
              \end{array}
 \right\}.\]
 It is clear that $p$ and $q$ are interchangeable in $\PP_2$. 
 In this case, both \sbass and \breakid detect (and break) this interchangeability. 
\end{example}

 \begin{example} 
 Consider the logic program 
  \[\PP_3=\left\{ \begin{array}{l}
               \lrule p\land q. \\
               0\leq \#\{p\} \leq 1.\\
               0\leq \#\{q\} \leq 1.
              \end{array}
 \right\}.\]
 It is clear that $p$ and $q$ are interchangeable in $\PP_3$. 
 \breakid detects this, while \sbass detects (and breaks) no symmetry. 
\end{example}

\begin{example}
Consider the logic program 
\[\PP_4=\left\{ \begin{array}{l}
               p\lor q \lrule p \land q.\\
               0\leq \#\{p\} \leq 1.\\
               0\leq \#\{q\} \leq 1.
              \end{array}
 \right\}.\]
 It is clear that $p$ and $q$ are interchangeable in $\PP_4$. 
 \breakid detects this, while \sbass detects (and breaks) no symmetry. 
 \end{example}
 
 \begin{example}\label{ex:facts}
Consider the logic program 
\[\PP_5=\left\{ \begin{array}{l}
               p.\\
               q.
              \end{array}
 \right\}.\]
 It is clear that $p$ and $q$ are interchangeable in $\PP_4$. 
 \breakid detects this, while \sbass detects no symmetry. 
 \end{example}
 In the last example, the fact that this symmetry is not detected does not have any practical consequences. Indeed, since these are only interchangeable facts, exploiting the symmetry will not help the solver. However, in examples such as Example \ref{ex:choice}, the difference is more important. If there are interchangeable atoms in choice rules, symmetry breaking can cut out exponentially large parts of the search space.

\subsection{Symmetry Breaking Constraints}
\newcommand{\lexvar}{\m{\prec_{\voc}}}
\newcommand{\lex}{\m{\prec}}
A classic approach to static symmetry breaking is to construct \emph{lex-leader constraints}. 
\begin{definition}[Lex-leader constraint~\cite{crawford1996symmetry}]
Let $\phi$ be a formula over $\voc$, $\pi$ a symmetry of $\phi$, $\lexvar$ an order on \voc and $\lex$ the induced lexicographic order on the set of interpretations over $\voc$.
A formula $LL_\pi$ over $\voc'\supseteq \voc$ is a \emph{lex-leader constraint} for $\pi$ if for each $\voc$-interpretation $I$, there exists a $\voc'$-extension of $I$ that satisfies $LL_\pi$ iff $I \lex \pi(I)$.
\end{definition}

In other words, each interpretation whose symmetric image under $\pi$ is smaller is invalidated by $LL_\pi$. It is easy to see that the conjunction of $LL_\pi$ for all $\pi$ in some $\Pi'\subseteq \Pi$ is a sound (but not necessarily complete) symmetry breaking constraint for $\Pi$.

An efficient encoding of the lex-leader constraint $LL_\pi$ as a conjunction of propositional clauses is given by Aloul et al.~\cite{2002Aloul}. Such clauses translate directly to ASP constraints. 
In order to make such an encoding, an order on $\voc$ needs to be chosen. 
The classic approach to symmetry breaking is to use the numbers as they occur in the numerical DIMACS CNF format or \lparse-\smodels intermediate format and to post for each symmetry in the set of generators a lex-leader constraint. 
This is the approach that is taken for instance in the SAT symmetry breaking tool \shatter  and in \sbass.

The drawback of this approach is discussed extensively in the work that introduced \breakid \mycite{breakid}. 
There, we show that this phase leaves room for several optimisations. 
\begin{compactitem}
 \item The set $\mathcal{G}$ of generators is often not good for symmetry breaking. Manipulating this set can result in symmetry breaking constraints with much stronger breaking power. 
 \item The order given by the numerical format is often suboptimal for symmetry breaking. The order should ``match'' the symmetries in $\mathcal{G}$. 
 \item If both of the above optimisations are executed, we effectively obtain a situation in which exponentially large subgroups of the group generated by $\mathcal{G}$ are broken \emph{completely} (without increasing the size of the symmetry breaking constraints). 
 \item Instead of breaking on the generators, one can generate a large number of very small (size 2) symmetry breaking clauses, resulting in stronger propagation power. 
 \item A slightly better encoding of the lex-leader constraints can be used \cite{sakallah2009symmetry}. 
\end{compactitem}

All of these optimisations are used in the ASP version of \breakid as well.


\section{Experiments}\label{sec:exp}
In this section, we experimentally compare \sbass and \breakid, using \clasp 3.1.4 as solver and \gringo 4.5.5 as grounder. 
We compare the number of solved instances for a set of four symmetric decision problems, and a set of two symmetric optimization problems from the 2013's ASP competition \cite{conf/lpnmr/AlvianoCCDDIKKOPPRRSSSWX13}.

\subsection{Setup}

The four decision problems in our benchmark set are \holes{}, \crew{}, \graceful{} and \nqueens{}. 

\holes{} is a set of 16 unsatisfiable pigeonhole instances where $n$ pigeons must be placed in $n-1$ different holes. $n$ takes values from $\{5,$\,\allowbreak{}$6,$\,\allowbreak{}$\ldots,$\,\allowbreak{}$14,$\,\allowbreak{}$15,$\,\allowbreak{}$20,$\,\allowbreak{}$30,$\,\allowbreak{}$50,$\,\allowbreak{}$70,$\,\allowbreak{}$100\}$.
The pigeons and holes are interchangeable, leading to a large symmetry group.

\crew{} is a set of 42 unsatisfiable airline crew scheduling instances, where optimality has to be proven for a minimal crew assignment given a moderately complex flight plan.
The instances are generated by hand, with the number of crew members ranging from $5$ to $25$.
Crew members have different attributes, but depending in the instance, multiple crew members exist with the same exact attribute set, making these crew members interchangeable.

\graceful{} consists of 60 satisfiable and unsatisfiable graceful graph instances, taken from 2013's ASP competition \cite{conf/lpnmr/AlvianoCCDDIKKOPPRRSSSWX13}. 
These instances require to label a graph's vertices and edges such that all vertices have a different label, all edges have a different label, and each edge's label is the difference of the labels of the vertices it connects.
The labels used are $\{0,1,\dots,n\}$, with $n$ the number of edges.
Any symmetry exhibited by the input graph is present, as well as a symmetry mapping each vertex' label $l$ to $n-l$.

\nqueens{} is a set of 4 large satisfiable N-Queens instances trying to fit $n$ queens on an $n$ by $n$ chessboard so that no queen threatens another. $n$ takes values from $\{50,100,150,200\}$.
The symmetries present in \nqueens{} are the rotational and reflective symmetries of the chessboard.

The two optimization problems are \valves{} and \still{}.
Both problems' models and instances are taken from 2013's ASP competition \cite{conf/lpnmr/AlvianoCCDDIKKOPPRRSSSWX13}, but manual symmetry breaking constraints were removed from the ASP specification.

\valves{} models connected pipelines in urban hydraulic networks, and features interchangeable valves.
Our instance set counts 50 instances.

\still{} models a fixpoint connected cell configuration in Conway's game of life. The game board exhibits rotational and reflective symmetry.
Our instance set counts 21 instances.

All experiments were run on on an Intel\textsuperscript{\textregistered} Xeon\textsuperscript{\textregistered} E3-1225 CPU with Ubuntu 14.04 Linux kernel 3.13 as operating system.
ASP specifications, instances, shell scripts and detailed experimental results are available online\footnote{\url{bitbucket.org/krr/breakid_asp}}.

The decision problems had 6GB RAM and 1000s timeout as resource limits, and the results exclude grounding time, as this is the same for each solving configuration, but include any time needed to detect symmetry and construct symmetry breaking constraints.
By default, \breakid{} limits the size of the symmetry breaking formula to 50 auxiliary variables for a given symmetry. To keep the comparison as fair as possible, \sbass was given the same limit.

The optimization problems had 8GB RAM and 50000s timeout as resource limits, and the results exclude identical grounding time and negligable symmetry breaking time.
No comparison with \sbass{} was made as \sbass{} does not support optimization statements.

\subsection{Decision problem results}
\begin{table}[t]
\centering
\begin{tabular}{ l | c | ccccc | ccccc | }			
  & \multicolumn{1}{l|}{\clasp} & \multicolumn{5}{c|}{\sbass}  & \multicolumn{5}{c|}{\breakid} \\
  & $\#$ & $\#$ & $t$ & $\pi$ & $r$ & $a$ & $\#$ & $t$ & $\pi$ & $r$ & $a$ \\
\hline
\holes{} (16) & 8 & 11 & 51.0 & 43.5 & 7729 & 1932 & \textbf{14} & 12.5 & 90.0 & 7383 & 2235 \\
\crew{} (42) & 32 & 36 & 0.0 & 7.8 & 734 & 183 & \textbf{41} & 0.0 & 100.8 & 654 & 184 \\
\graceful{} (60) & \textbf{33} & 28 & 0.7 & 5.5 & 1085 & 271 & 32 & 2.1 & 102.7 & 1161 & 347 \\
\nqueens{} (4) & 4 & 4 & 33.7 & 2.0 & 392 & 98 & 4 & 41.1 & 126.0 & 426 & 100 \\
\end{tabular}
\caption{Experimental results of (i) \clasp without symmetry breaking preprocessor, (ii) with \sbass, and (iii) with \breakid. $\#$ represents the number of solved instances within resource bounds, $t$ the average symmetry preprocessing time in seconds (including detection and construction of symmetry breaking constraints), $\pi$ the average number of symmetry generators detected by \saucy, $r$ the average number of symmetry breaking rules constructed, and $a$ the average number of auxiliary atoms introduced.}%
\label{tbl:results}
\end{table}%
Table~\ref{tbl:results} summarizes the results of the decision problems.
Analyzing the results on \holes{}, \clasp gets lost in symmetric parts of the search tree, solving only $8$ instances (up to $12$ pigeons).
\sbass{} can only solve three more instances (up to $15$ pigeons), as the derived symmetry generators do not suffice to construct strong symmetry breaking constraints.
These results are consistent with the results of Drescher et al.~\mycite{sbass}. 
\breakid{} detects more structure, solving all but two instances (the largest being solved contains $50$ pigeons).
As far as symmetry preprocessing time goes, both \sbass and \breakid spend a significant time detecting symmetry, especially on the larger instances with more than 20 pigeons. 
To our surprise, \breakid requires only a quarter of the symmetry preprocessing time \sbass uses. This difference is entirely due to \saucy{} needing much more time to detect automorphisms on the graph encoding of \sbass{} than on the graph encoding of \breakid{}. 
Recall that, compared to \breakid{}, \sbass{} has the smaller encoding graph for automorphism detection due to the use of directed edges.
Hence, the most plausible hypothesis to explain the difference in symmetry detection time is simply that \saucy{} is sensitive to differences in graph encodings, which gets magnified by large problem instances.
Nonetheless, even for the largest instance with 100 pigeons, symmetry preprocessing by \sbass{} did not reach the timeout limit.

The results on \crew{} are similar to \holes{}: \breakid outperforms \sbass, which in turn outperforms plain \clasp. On the other hand, symmetry preprocessing time is negligible for \crew{}. This is mainly due to the sizes of the ground programs remaining relatively small; less than 3000 rules for the largest ground program in the \crew{} instance set.

Continuing with \graceful{}, it is striking that the number of solved instances is \emph{reduced} by symmetry breaking.
Upon closer inspection, this is only the case for satisfiable instances.
For unsatisfiable \graceful{} instances, \sbass{} and \breakid{} both solve four instances, two more than \clasp{}.
This discrepancy is not uncommon, as static symmetry breaking reduces the search space by sometimes removing easy-to-find solutions.
These results are also consistent with those reported by Drescher et al.~\mycite{sbass}. 
Focusing on symmetry preprocessing time, \sbass{} is faster than \breakid{}.
This is consistent with  our previous findings~\mycite{breakid}, where we argued that deriving better symmetry breaking clauses incurs extra overhead.

Lastly, for \nqueens{}, all approaches solve all four instances easily. Again \sbass{} is faster than \breakid{}, which is explained by the same reason as for \graceful{}. However, \breakid{}'s preprocessing time remains well within timeout bounds.

It is also interesting to note that on all problem families \sbass and \breakid have about the same number of symmetry breaking constraints and auxiliary variables introduced, incurring roughly the same memory overhead for both approaches.
However the number of symmetry generators detected by \breakid is much larger.
Upon closer inspection, it turns out that \breakid often detects (trivial) interchangeability of \emph{facts} in the program. There is no advantage in breaking such symmetry, but it does not hurt performance either.
We are not sure why \sbass does not detect this kind of symmetries. Either it is because of the directional encoding, resulting essentially in a bug similar to Example \ref{ex:facts}, or it is because of an optimisation for facts we have no knowledge of. 

We conclude that on the decision problem benchmark set, \breakid outperforms \sbass, especially for problems with interchangeable objects such as \holes{} and \crew{}. The price to be paid is a bit more symmetry preprocessing overhead, though the size of the symmetry breaking formula remains comparable between both approaches. This is to be expected, as \breakid{} compared similarly to \shatter{} for SAT problems.

\subsection{Optimization problem results}
\begin{figure}[bt]
    \centering
    \includegraphics[width=1.01\textwidth]{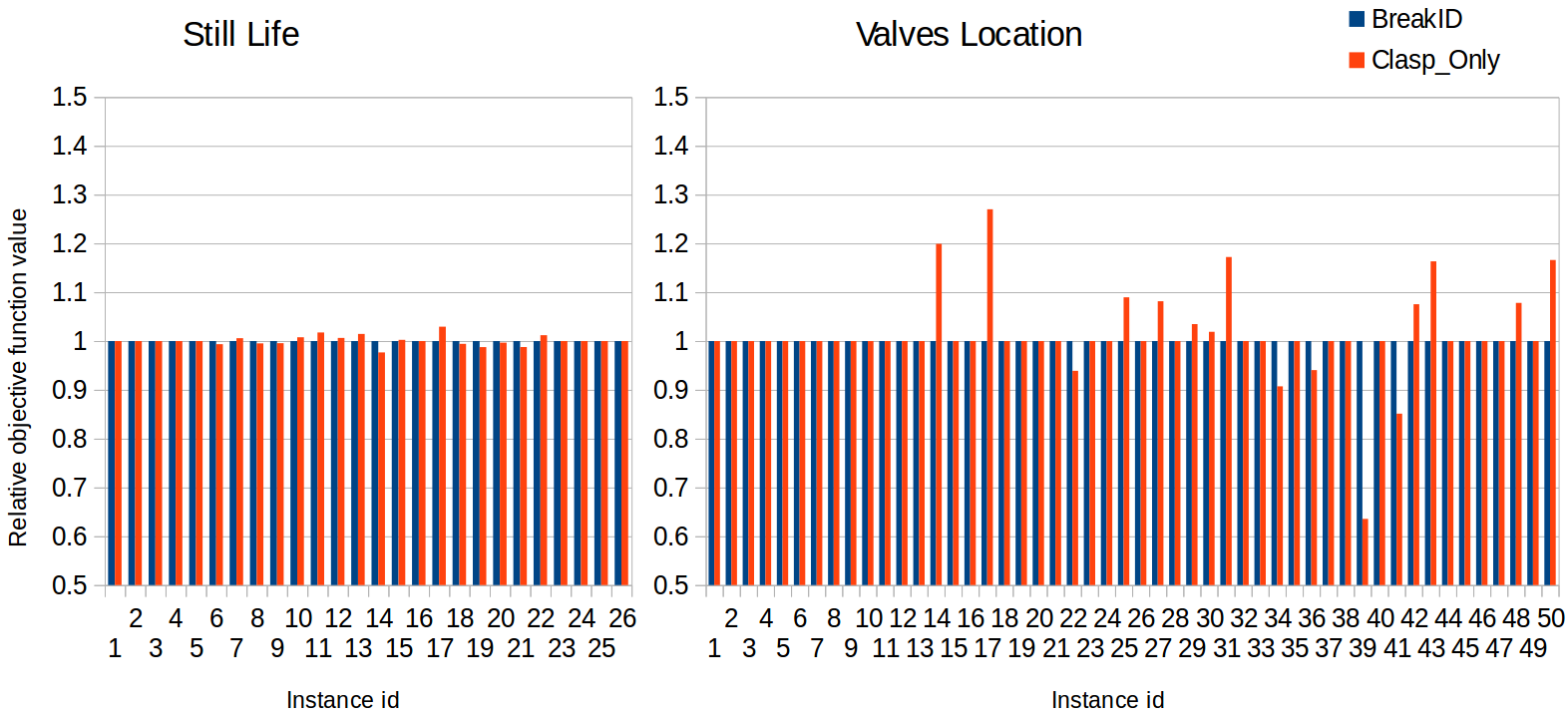}
    \caption{The relative objective value of the best solution found after 50000s of search time for \still{} and \valves{}. Lower is better.}
    \label{fig:still_valves}
\end{figure}
Figure \ref{fig:still_valves} shows the relative objective value of the best solution found after 50000s of search time for \still{} and \valves{}.
Even though \breakid{} detects and breaks significant symmetry for both problems, and often infers interchangeability in \valves{}, the resulting symmetry breaking constraints do not seem to vastly improve the final objective value for these two benchmark families.
For \still{}, the resulting objective value is virtually identical.
For \valves{}, the objective value for \breakid{}'s run is improved for 11 instances, while it has worsened for 5 instances, giving a small edge to \breakid{} compared to \clasp{}.
When looking at the number of instances for which optimality was proven, both approaches were able to prove optimality for 5 \still{} instances and 15 \valves{} instances.

We conclude that for optimization problems \still{} and \valves{}, our symmetry breaking approach detects and breaks symmetry, but any resulting speedups are small at best.

\section{Discussion and Conclusion}\label{sec:concl}
In this paper, we describe our system \breakid, which performs \emph{static}, \emph{ground-level} symmetry breaking for ASP. 
This type of symmetry breaker is the ``easiest'' type to implement. It serves merely as a preprocessor for the solver. Other types of symmetry breaking exist as well. 
For instance, in SAT and constraint programming, \emph{dynamic} symmetry breaking is well-studied \cite{Schaafsma2009,cspsat/DevriendtBB14,symchaff}. The disadvantage of these approaches is that they require modifying the internals of the solver at hand. The advantage, however, is that such approaches are often more memory-friendly, since they do not require to create all the symmetry breaking constraints prior to search, are not dependent on any fixed variable order chosen a priori, and do not remove easy-to-find solutions from the search space. 
Another line of research in symmetry breaking is \emph{first-order symmetry detection}, where symmetries are detected on the predicate level. In constraint programming, work has been done on composing the symmetry properties of global constraints into a symmetry group of the original problem \cite{VanHentenryck:2005:CDS}. However, most existing symmetry detection implementations also detect symmetry on the ground level. 

Recently, we proposed first-order symmetry detection and breaking algorithms for the \idp system \mycite{IDP}, a system whose input language relates closely to ASP. The main advantage of such an approach is that it allows to use the \emph{structure} of the program to detect certain kinds of symmetries without the need for an automorphism tool. 
The drawbacks are that they cannot detect all types of symmetry and that they are required to interpret a much richer (first-order) language. Thus, such symmetry detection requires a significantly larger implementation effort. 
We believe it would definitely benefit the community to have a version of \gringo, or another ASP grounder, with support for first-order symmetry detection. 

The techniques we presented here are not bound to the stable semantics. They are equally applicable for instance to the well-founded semantics \mycite{wfs}, and hence apply as well to the logic \foid \mycite{fodot} (which combines logic programs under the well-founded semantics with propositional theories). The reason is that our techniques detect \emph{syntactic} symmetries and hence are agnostic about the semantics at hand. 

We presented experimental results comparing our tool to the existing ASP symmetry breaker \sbass.
For the employed decision problem benchmark set, \breakid outperforms \sbass, with a small drawback in the form of greater symmetry preprocessing overhead.
\breakid also detects and breaks symmetry on optimization problems, though significant speedups have yet to be observed.

The result of this work is an easy-to-use tool that can significantly improve performance on non-optimised encodings and works with all modern ASP solvers. It can be downloaded at \url{bitbucket.org/krr/breakid}. 
Our expectation is that using this tool yields no (or very little) performance gain on highly optimised encodings, crafted by experts. 
However, for average users of ASP tools, \breakid can cut away large parts of the search space and is an improvement to \sbass, both from an applicability as well as an efficiency point-of-view.

\section*{Acknowledgements}
This research was supported by the project GOA 13/010 Research Fund
KU Leuven and projects G.0489.10, G.0357.12 and G.0922.13 of FWO
(Research Foundation - Flanders). 
Bart Bogaerts is supported by the Finnish Center of Excellence in Computational
Inference Research (COIN) funded by the Academy of Finland (grant \#251170).
%

\bibliographystyle{splncs03}
\bibliography{idp-latex/krrlib}

\end{document}